\documentclass[11pt]{article}
\usepackage{coling2018}
\usepackage{times}
\usepackage{url}
\usepackage{latexsym}
\usepackage{graphicx}
\usepackage{amsmath}
\usepackage{amsfonts}
\usepackage{array}
\usepackage{tabu}

\title{Deep Enhanced Representation for\\Implicit Discourse Relation Recognition}

\author{Hongxiao Bai$^{1,2}$, Hai Zhao$^{1,2,}$\thanks{
  \,\, Corresponding author.
  This paper was partially supported by
  National Key Research and Development Program of China (No. 2017YFB0304100),
  National Natural Science Foundation of China (No. 61672343 and No. 61733011),
  Key Project of National Society Science Foundation of China (No. 15-ZDA041),
  The Art and Science Interdisciplinary Funds of Shanghai Jiao Tong University (No. 14JCRZ04).
  }  \\
$^1$Department of Computer Science and Engineering,\\
Shanghai Jiao Tong University, Shanghai, China \\
$^2$Key Laboratory of Shanghai Education Commission for Intelligent Interaction\\
and Cognitive Engineering, Shanghai Jiao Tong University, Shanghai, China\\
  {\tt baippa@sjtu.edu.cn, zhaohai@cs.sjtu.edu.cn} \\}

\date{}

\begin{document}
\maketitle
\begin{abstract}
  Implicit discourse relation recognition is a challenging task as the relation prediction without explicit connectives
  in discourse parsing needs understanding of text spans and
  cannot be easily derived from
  surface features from the input sentence pairs.
  Thus, properly representing the text is very crucial to this task.
  In this paper, we propose a model augmented with different grained text representations,
  including character, subword, word, sentence, and sentence pair levels.
  The proposed deeper model is evaluated on the benchmark treebank and achieves
  state-of-the-art accuracy with greater than 48\% in 11-way and $F_1$ score greater than 50\% in 4-way classifications
  for the first time according to our best knowledge.
\end{abstract}

\section{Introduction}

\blfootnote{
    %
    %
    \hspace{-0.65cm}  
    This work is licenced under a Creative Commons Attribution 4.0 International Licence.
    Licence details: \url{http://creativecommons.org/licenses/by/4.0/}    %
    %
    %
    %
}

Discourse parsing is a fundamental task in natural language processing (NLP) which 
determines the structure of the whole discourse and identifies the relations between discourse spans
such as clauses and sentences.
Improving this task can be helpful to many downstream tasks such as
machine translation~\cite{li-carpuat-nenkova:2014:P14-2},
question answering~\cite{jansen2014discourse}, and so on.
As one of the important parts of discourse parsing, implicit discourse
relation recognition task is to find the relation between two spans
without explicit connectives (e.g., \emph{but, so}, etc.),
and needs recovering the relation from semantic understanding of texts.

The Penn Discourse Treebank 2.0 (PDTB 2.0)~\cite{prasad2008penn} is a benchmark corpus for discourse
relations. In PDTB style, the connectives can be explicit or implicit, and one entry of the data is
separated into \emph{Arg1} and \emph{Arg2}, accompanied with a relation sense.
Since the release of PDTB 2.0 dataset, many methods have been
proposed, ranging from traditional feature-based methods
~\cite{lin-kan-ng:2009:EMNLP,pitler-louis-nenkova:2009:ACLIJCNLP}
to latest neural-based methods~\cite{qin-EtAl:2017:Long,lan-EtAl:2017:EMNLP20172}.
Especially through many neural network methods used for this task such as
convolutional neural network~(CNN)~\cite{qin2016conll},
recursive neural network~\cite{TACL536}, embedding improvement~\cite{wu-EtAl:2017:Short},
attention mechanism~\cite{liu-li:2016:EMNLP2016}, gate mechanism~\cite{chen-EtAl:2016:P16-13},
multi-task method~\cite{lan-EtAl:2017:EMNLP20172},
the performance of this task has improved a lot since it was first introduced.
However, this task is still very challenging with the highest reported accuracy still
lower than 50\% due to the hardness for the machines to understand the text meaning and the relatively
small task corpus.

In this work, we focus on improving the learned representations of sentence pairs
to address the implicit discourse relation recognition.
It is well known that text representation is the core part of state-of-the-art
deep learning methods for NLP tasks, and improving the representation from all perspective
will benefit the concerned task.

The representation is improved by two ways in
our model through three-level hierarchy.
The first way is embedding augmentation.
Only with informative embeddings, can the final representations be better.
This is implemented in our word-level module.
We augment word embeddings with subword-level embeddings and pre-trained ELMo embeddings.
Subwords coming from unsupervised segmentation demonstrate a better consequent performance
than characters for being a better minimal representation unit.
The pre-trained contextualized word embeddings (ELMo) can make the embeddings
contain more contextual information which is also involved with character-level inputs.
The second way is a deep residual bi-attention encoder.
Since this task is about classifying sentence pairs, the encoder is implemented in
sentence and sentence-pair levels.
A deeper model can support richer representations but is hard to train, especially with
a small dataset. So we apply residual connections~\cite{HeZRS16} to each module
for facilitating signal propagation and alleviating gradient degradation.
The stacked encoder blocks make the single sentence representation richer
and bi-attention module mixes two sentence representations focusingly.
With introducing richer and deeper representation enhancement,
we report the deepest model so far for the task.

Our representation enhanced model will be evaluated on the benchmark PDTB 2.0
and demonstrate state-of-the-art performance to verify its effectiveness.

This paper is organized as follows. Section 2 reviews related work. Section 3 introduces our model.
Section 4 shows our experiments and analyses the results. Section 5 concludes this work.

\section{Related Work}

After the release of Penn Discourse Treebank 2.0, many works have been made to solve this concerned task.
\newcite{lin-kan-ng:2009:EMNLP} is the first work who considered the second-level classification
of the task by empirically evaluating the impact of surface features.
Feature based methods~\cite{pitler-louis-nenkova:2009:ACLIJCNLP,zhou2010predicting,chen-wang-zhao:2015:CoNLL-ST,li2016conll}
mainly focused on using linguistic, or semantic features
from the discourse units, or the relations between unit pairs and word pairs.
\newcite{zhang-EtAl:2015:EMNLP4}
is the first one who modeled this task using end-to-end neural network and gained
great performance improvement.
Neural network methods also used by lots of works~\cite{zhang2016probabilistic,P16-1039} for better performance.
Since then, a lot of methods have been proposed. \newcite{braud2015comparing} found that
word embeddings trained by neural networks is very useful to this task.
\newcite{qin-zhang-zhao:2016:COLING} augmented their system with character-level and
contextualized embeddings.
Recurrent networks and convolutional networks have been used as basic blocks in many works~
\cite{ji-haffari-eisenstein:2016:N16-1,ronnqvist-schenk-chiarcos:2017:Short,qin2016conll}.
\newcite{TACL536} used recursive neural networks. Attention mechanism was used by
\newcite{liu-li:2016:EMNLP2016}, \newcite{cai2017discourse} and others.
\newcite{wu-EtAl:2016:EMNLP2016} and \newcite{lan-EtAl:2017:EMNLP20172} applied
multi-task component. \newcite{qin-EtAl:2017:Long} utilized adversarial nets
to migrate the connective-based features to implicit ones.

Sentence representation is a key component in many NLP tasks. Usually, better representation means
better performance. Plenty of work on language modeling has been done, as language modeling can supply
better sentence representations. Since the pioneering work of \newcite{Bengio2006},
neural language models have been well developed~
\cite{MikolovZ12,WilliamsPMAR15,kim2016character}.
Sentence representation is directly handled in a series of work.
\newcite{lin2017structured} used self attention mechanism and used matrix to represent sentence,
and \newcite{conneau-EtAl:2017:EMNLP2017} used encoders pre-trained on
SNLI~\cite{bowman-EtAl:2015:EMNLP} and MultiNLI~\cite{williams2017broad}.

Different from all the existing work, for the first time to our best knowledge,
this work is devoted to an empirical study on
different levels of representation enhancement for implicit discourse relation classification task. 

\section{Model}

\subsection{Overview}

\begin{figure}[ht]
  \centering
  \includegraphics[width=1\textwidth]{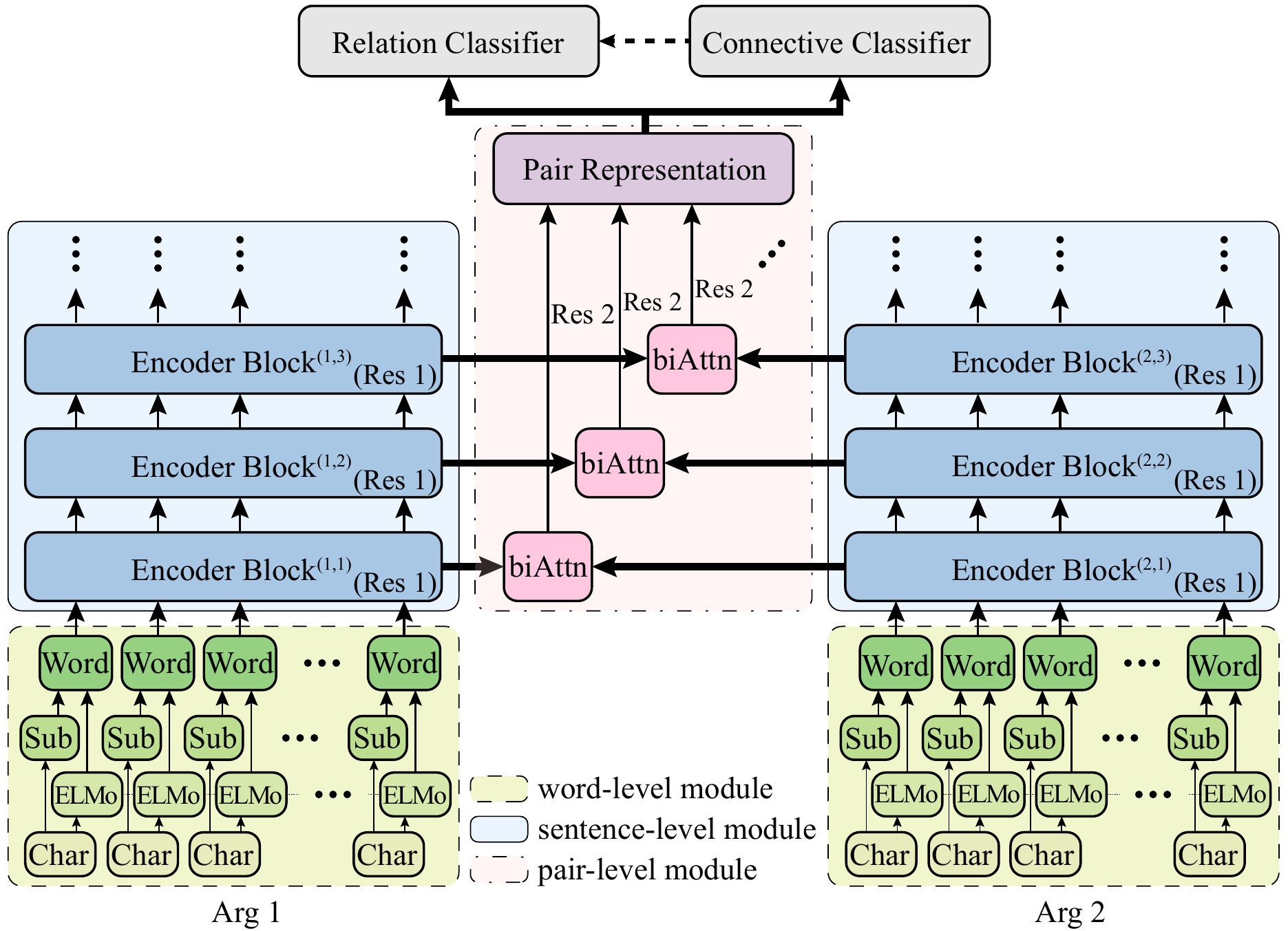}
  \caption{Model overview.}
  \label{fig:model}
\end{figure}

Figure~\ref{fig:model} illustrates an overview of our model, which is mainly
consisted of three parts: word-level module, sentence-level module, and pair-level module.
Token sequences of sentence pairs~(\emph{Arg1} and \emph{Arg2}) are encoded by word-level module first
and every token becomes a word embedding augmented by subword and ELMo.
Then these embeddings are fed to sentence-level module and processed by stacked encoder
blocks~(CNN or RNN encoder block).
Every block layer outputs representation for each token.
Furthermore, the output of each layer is processed by bi-attention module
in the pair-level module, and
concatenated to pair representation, which is finally sent to classifiers
which are multiple layer perceptrons~(MLP) with softmax. 
The model details are given in the rest of this section.

\subsection{Word-Level Module}

An inputed token sequence of length $N$ is encoded by the word-level module into an
embedding sequence $(\mathbf{e}_1, \mathbf{e}_2, \mathbf{e}_3, \cdots, \mathbf{e}_N)$.
For each embedded token $\mathbf{e}_i$, it is concatenated from three parts,
\begin{equation}
  \mathbf{e}_i = [\mathbf{e}_i^w;~ \mathbf{e}_i^s;~ \mathbf{e}_i^c] \in \mathbb{R}^{d_e}
\label{emb}\end{equation}
$\mathbf{e}_i^w \in \mathbb{R}^{d_w}$ is pre-trained word embedding for this token, and is fixed during the
training procedure.
Our experiments show that fine-tuning the embeddings slowed down the training without better performance.
$\mathbf{e}_i^s \in \mathbb{R}^{d_s}$ is subword-level embedding encoded by subword encoder.
$\mathbf{e}_i^c \in \mathbb{R}^{d_c}$ is contextualized word embedding encoded by pre-trained ELMo encoders,
whose parameters are also fixed during training.
Subword is merged from single-character segmentation and the input of ELMo encoder is also character.

\subsubsection*{Subword Encoder}

Character-level embeddings have been used widely in lots of works and its effectiveness
is verified for out-of-vocabulary~(OOV) or rare word representation.
However, character is not a natural minimal unit for there exists word
internal structure, we thus introduce a subword-level embedding instead.

Subword units can be computationally discovered by unsupervised segmentation over words
that are regarded as character sequences.
We adopt byte pair encoding (BPE) algorithm introduced by
\newcite{sennrich-haddow-birch:2016:P16-12}
for this segmentation.
BPE segmentation actually relies on a series of iterative merging operation over
bigrams with the highest frequency.
The number of merging operation times is roughly equal to the result subword
vocabulary size.

\begin{figure}[ht]
  \centering
  \includegraphics[width=0.4\textwidth]{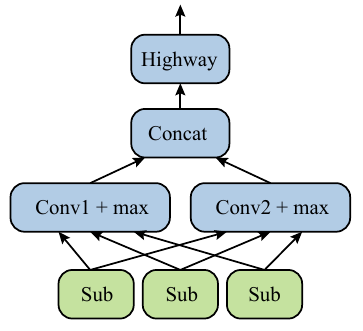}
  \caption{Subword encoder.}
  \label{fig:sub}
\end{figure}

For each word, the subword-level embedding is encoded by a subword encoder
as in Figure~\ref{fig:sub}.
Firstly, the subword sequence~(of length $n$) of the word is mapped to subword embedding
sequence $(\mathbf{se}_1, \mathbf{se}_2, \mathbf{se}_3, \cdots, \mathbf{se}_n)$ (after padding),
which is randomly initialized.
Then $K$~(we empirically set $K$=$2$) convolutional operations $Conv_1, Conv_2, \cdots, Conv_K$
followed by max pooling operation are applied to the embedding sequence,
and the sequence is padded before the convolutional operation.
For the $i$-th convolution kernel $Conv_i$,
suppose the kernel size is $k_i$,
then the output of $Conv_i$ on embeddings $\mathbf{se}_{j}$ to $\mathbf{se}_{j+k_i-1}$ is
\[\begin{split}
  \mathbf{C}_j &= \tanh \Big(Conv_i[\mathbf{se}_{j}: \mathbf{se}_{j+k_i-1}] \Big)
\end{split}\]
The final output of $Conv_i$ after max pooling is
\[\begin{split}
  \mathbf{u}_i &= \mathop{maxpool}{(\mathbf{C}_1,~ \cdots,~ \mathbf{C}_j,~ \cdots,~ \mathbf{C}_n)}
\end{split}\]
Finally, the $K$ outputs are concatenated,
\[
  \mathbf{u} = [\mathbf{u}_1;~ \mathbf{u}_2;~ \cdots;~ \mathbf{u}_K] \in \mathbb{R}^{d_s}
\]
to feed a highway network~\cite{highway},
\begin{eqnarray}
  \mathbf{g} &=& \sigma(\mathbf{W}_g \mathbf{u}^T + \mathbf{b}_g) \in \mathbb{R}^{d_s} \nonumber \\
  \mathbf{e}_i^s &=& \mathbf{g} \odot \mathop{ReLU}(\mathbf{W}_h \mathbf{u}^T + \mathbf{b}_h)
  + (\mathbf{1} - \mathbf{g}) \odot \mathbf{u} \nonumber \\
  &\in& \mathbb{R}^{d_s}
\label{es}\end{eqnarray}
where $\mathbf{g}$ denotes the gate,
and $\mathbf{W}_g \in \mathbb{R}^{d_s \times d_s}, \mathbf{b}_g \in \mathbb{R}^{d_s},
\mathbf{W}_h \in \mathbb{R}^{d_s \times d_s}, \mathbf{b}_h \in \mathbb{R}^{d_s}$ are parameters.
$\odot$ is element-wise multiplication.
The above Eq.~\ref{es} gives the subword-level embedding for the $i$-th word.

\subsubsection*{ELMo}
ELMo~(Embeddings from Language Models)~\cite{Peters2018ELMo} is a pre-trained contextualized
word embeddings involving character-level representation.
It is shown useful in some works~\cite{he_srl_acl2018,lee_coref_naacl2018}.
This embedding is trained by bidirectional language models on large corpus using character sequence
for each word token as input.
The ELMo encoder employs CNN and highway networks over characters,
whose output is given to a multiple-layer biLSTM with residual connections.
Then the output is contextualized embeddings for each word.
It is also can be seen as a hybrid encoder for character, word, and sentence.
This encoder can add lots of contextual information to each word, and can
ease the semantics learning of the model.

For the pre-trained ELMo encoder, the output is the result of the last two biLSTM layers.
Suppose $\mathbf{c}_i$ is the character sequence of $i$-th word in a sentence, then the
encoder output is
\[
  [\cdots, \mathbf{h}_i^0, \cdots;~ \cdots, \mathbf{h}_i^1, \cdots]
  = \mathop{ELMo}(\cdots, \mathbf{c}_i, \cdots)
\]
where $\mathbf{h}_i^0$ and $\mathbf{h}_i^1$ denote the outputs of first and second layers of ELMo encoder for $i$-th word.

Following \newcite{Peters2018ELMo}, we use
a self-adjusted weighted average of $\mathbf{h}_i^0, \mathbf{h}_i^1$,
\[\begin{split}
  \mathbf{s} &= \mathop{softmax}(\mathbf{w}) \in \mathbb{R}^2\\
  \mathbf{h} &= \gamma \sum_{j=0}^1 s_j \mathbf{h}_i^j \in \mathbb{R}^{d_c^{\prime}}
\end{split}\]
where $\gamma \in \mathbb{R}$ and $\mathbf{w} \in \mathbb{R}^2$ are parameters tuned during training
and $d_c^{\prime}$ is the dimension of the ELMo encoder's outputs.
Then the result is fed to a feed forward network to reduce its dimension,
\begin{equation}
  \mathbf{e}_i^c = \mathbf{W}_c \mathbf{h}^T + \mathbf{b}_c \in \mathbb{R}^{d_c}
\label{ec}\end{equation}
$\mathbf{W}_c \in \mathbb{R}^{d_c^{\prime} \times d_c}$ and $\mathbf{b}_c \in \mathbb{R}^{d_c}$ are parameters.
The above Eq.~\ref{ec} gives ELMo embedding for the $i$-th word.

\subsection{Sentence-Level Module}

The resulting word embeddings $\mathbf{e}_i$~(Eq.~\ref{emb}) are sent to
sentence-level module.
The sentence-level module is composed of stacked encoder blocks.
The block in each layer receives output of the previous layer as input and sends
output to next layer. It also sends its output to the pair-level module.
Parameters in different layers are not the same.

We consider two encoder types, convolutional type and recurrent type.
We only use one encoder type in one experiment.

For the sentence-level module for different arguments~(\emph{Arg1} and \emph{Arg2}),
many previous works used same parameters to encode different arguments, that is,
one encoder for two type arguments.
But as indicated by \newcite{prasad2008penn}, \emph{Arg1} and \emph{Arg2} may have
different semantic perspective, we thus introduce argument-aware parameter settings
for different arguments.

\begin{figure}[ht]
  \centering
  \begin{minipage}[t]{0.48\textwidth}
  \centering
  \includegraphics[width=0.95\textwidth]{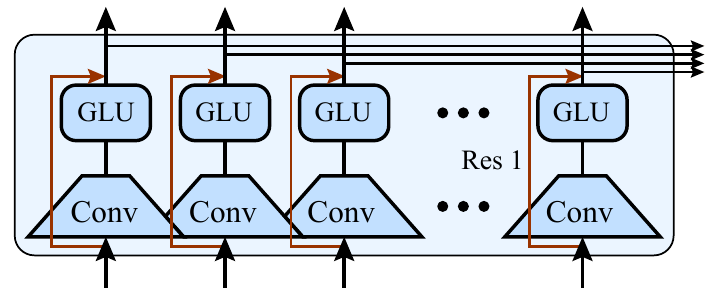}
  \caption{Convolutional encoder block.}
  \label{fig:convb}
  \end{minipage}
  \hfill
  \begin{minipage}[t]{0.48\textwidth}
  \centering
  \includegraphics[width=0.95\textwidth]{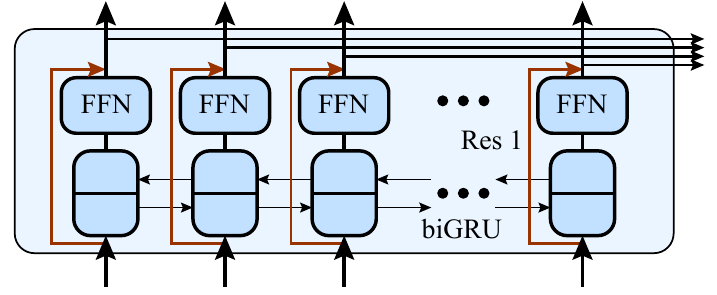}
  \caption{Recurrent encoder block.}
  \label{fig:rnnb}
  \end{minipage}
\end{figure}

\subsubsection*{Convolutional Encoder Block}

Figure~\ref{fig:convb} is the convolutional encoder block.
Suppose the input for the encoder block is $\mathbf{x}_i ~ (i=1, \cdots, N)$, then $\mathbf{x}_i \in \mathbb{R}^{d_e}$.
The input is sent to a convolutional layer and mapped to output
$\mathbf{y}_i = [\mathbf{A}_i \; \mathbf{B}_i] \in \mathbb{R}^{2d_e}$.
After the convolutional operation, gated linear units~(GLU)~\cite{dauphin2016language}
is applied, i.e.,
\[
  \mathbf{z}_i = \mathbf{A}_i \odot \sigma(\mathbf{B}_i) \in \mathbb{R}^{d_e}
\]
There is also a residual connection~(Res 1) in the block,
which means adding the output of $\mathop{GLU}$ and the input of the block as
final output,
so $\mathbf{z}_i + \mathbf{x}_i$
is the output of the block corresponding to the input $\mathbf{x}_i$.
The output $\mathbf{z}_i + \mathbf{x}_i$ for all $i = 1, \cdots, N$
is sent to both the next layer and the pair-level module as input.

\subsubsection*{Recurrent Encoder Block}

Similar to the convolutional one, recurrent encoder block is shown in Figure~\ref{fig:rnnb}.
The input $\mathbf{x}_i$ is encoded by a biGRU~\cite{cho-EtAl:2014:EMNLP2014} layer first,
\[
  \mathbf{y}_i = \mathop{biGRU}(\mathbf{x}_i) \in \mathbb{R}^{2d_e}
\]
then this is sent to a feed forword network,
\begin{equation}
  \mathbf{z}_i = \mathbf{W}_r \mathbf{y}_i^T + \mathbf{b}_r \in \mathbb{R}^{d_e}
\label{ffn}\end{equation}
$\mathbf{W}_r \in \mathbb{R}^{2d_e \times d_e}$ and $\mathbf{b}_r \in \mathbb{R}^{d_e}$ are parameters.
There is also a similar residual connection (Res 1) in the block,
so $\mathbf{z}_i + \mathbf{x}_i$ for all $i = 1, \cdots, N$
is the final output of the recurrent encoder block.

\subsection{Pair-Level Module}
Through the sentence-level module, the word representations are contextualized,
and these contextualized representations of each layer are sent to
pair-level module.

Suppose the encoder block layer number is $l$,
and the outputs of $j$-th block layer for
\emph{Arg1} and \emph{Arg2} are $\mathbf{v}_1^j, \mathbf{v}_2^j \in \mathbb{R}^{N \times d_e}$,
each row of which is the embedding for the corresponding word.
$N$ is the length of word sequence~(sentence).
Each sentence is padded or truncated to let all sentences have the same length.
They are sent to a bi-attention module, the attention matrix is
\[
  \mathbf{M}_j = (\mathop{FFN}(\mathbf{v}_1^j)) {\mathbf{v}_2^j}^T
  \in \mathbb{R}^{N \times N}
\]
$\mathop{FFN}$ is a feed froward network (similar to Eq.~\ref{ffn}) applied to
the last dimension corresponding to the word.
Then the projected representations are
\[\begin{split}
  \mathbf{w}_2^j &= \mathop{softmax}(\mathbf{M}_j) {\mathbf{v}_2^j} \in \mathbb{R}^{N \times d_e}\\
  \mathbf{w}_1^j &= \mathop{softmax}(\mathbf{M}_j^T) {\mathbf{v}_1^j} \in \mathbb{R}^{N \times d_e}
\end{split}\]
where the $\mathop{softmax}$ is applied to each row of the matrix.
We apply 2-max pooling on each projected representation and concatenate them as output
of the $j$-th bi-attention module
\[
  \mathbf{o}_j = [\mathop{top2}(\mathbf{w}_1^j);~ \mathop{top2}(\mathbf{w}_2^j)]
  \in \mathbb{R}^{4 d_e}
\]
The number of max pooling operation (top-2) is selected from experiments and it is
a balance of more salient features and less noise.
The final pair representation is
\begin{equation}
  \mathbf{o} = [\mathbf{o}_1, \mathbf{o}_2, \cdots, \mathbf{o}_l] \in \mathbb{R}^{4 l d_e}
\label{paireq}\end{equation}
Since the output is concatenated from different layers
and the outputs of lower layers are sent directly to the final representation,
this also can be seen as residual connections~(Res 2).
Then the output as Eq.~\ref{paireq} is fed to an MLP classifier with softmax.
The parameters for bi-attention modules in different levels are shared.

\subsection{Classifier}
We use two classifiers in our model.
One is for relation classification, and another one is for connective classification.
The classifier is only a multiple layer perceptron (MLP) with softmax layer.
\newcite{qin-EtAl:2017:Long} used adversarial method to utilize the connectives,
but this method is not suitable for our adopted attention module
since the attended part of a sentence will be distinctly different when
the argument is with and without connectives.
They also proposed a multi-task method that augments the model with
an additional classifier for connective prediction, and the input of it is also the pair representation.
It is straightforward and simple enough, and can help the model learn better representations,
so we include this module in our model.
The implicit connectives are provided by PDTB 2.0 dataset,
and the connective classifier is only used during training.
The loss function for both classifiers is cross entropy loss,
and the total loss is the sum of the two losses, i.e.,
$Loss = Loss_{relation} + Loss_{connective}$.

\section{Experiments\footnote{The code for this paper is available at \url{https://github.com/diccooo/Deep_Enhanced_Repr_for_IDRR}}}

Our model is evaluated on the benchmark PDTB 2.0 for two types of classification tasks.

PDTB 2.0 has three levels of senses: Level-1 \emph{Class}, Level-2 \emph{Type},
and Level-3 \emph{Subtypes}.
The first level consists of four major relation Classes:
COMPARISON, CONTINGENCY, EXPANSION, and TEMPORAL. The second level contains 16 Types.

All our experiments are implemented by PyTorch\footnote{\url{http://pytorch.org/}}.
The pre-trained ELMo encoder is from AllenNLP toolkit~\cite{allennlp}
\footnote{\url{http://allennlp.org/}}.

\subsection{11-way Classification}

\subsubsection*{Dataset Setup}

Following the settings of \newcite{qin-EtAl:2017:Long}, we use two splitting methods
of PDTB dataset for comprehensive comparison.
The first is PDTB-Lin~\cite{lin-kan-ng:2009:EMNLP}, which uses section
2-21, 22 and 23 as training, dev and test sets respectively.
The second is PDTB-Ji~\cite{TACL536}, which uses section
2-20, 0-1, and 21-22 as training, dev and test sets respectively.
According to \newcite{TACL536}, five relation types have few training instances and
no dev and test instance. Removing the five types, there remain 11 second level types.
During training, instances with more than one annotated relation types are considered
as multiple instances, each of which has one of the annotations.
At test time, a prediction that matches one of the gold types is considered as correct.
All sentences in the dataset are padded or truncated to keep the same 100-word length.

\subsubsection*{Model Details}

For the results of both splitting methods, we share some hyperparameters.
Table~\ref{table:hyper1} is some of the shared hyperparameter settings.
The pre-trained word embeddings are 300-dim \emph{word2vec}~\cite{mikolov2013distributed} pre-trained
from Google News\footnote{\url{https://code.google.com/archive/p/word2vec/}}.
So $d_w = 300, d_s = 100, d_c = 300$,
then for the final embedding ($\mathbf{e}_i$), $d_e = 700$.
For the encoder block in sentence-level module,
kernel size is same for every layer.
We use AdaGrad optimization~\cite{duchi2011adaptive}.

The encoder block layer number is different for the two splitting methods.
The layer number for PDTB-Ji splitting method is 4,
and the layer number for PDTB-Lin splitting method is 5.

\begin{table}[ht]
  \centering
  \begin{tabu}{cc|cc}
    \tabucline[0.65pt]{-}
  Hyperparameter & Value & Hyperparameter & Value \\ \hline
  BPE merge operation num. & $1k$ & classifier layer num. & 1 \\
  subword embedding dim.~($d_s$) & 50 & embedding dropout & 0.4 \\
  subword CNN kernel num. & 2 & encoder block dropout & 0.4 \\
  subword CNN kernel size & [2, 3] & classifier dropout & 0.3 \\
  reduced ELMo embedding dim. & 300 & learning rate & 0.001 \\
  encoder block type & Conv. & batch size & 64 \\
  encoder block kernel size & 5 \\
    \tabucline[0.65pt]{-}
  \end{tabu}
  \caption{Shared hyperparameter settings. Before dimension reducing, the dimension of pre-trained ELMo embedding is 1024.}
  \label{table:hyper1}
\end{table}
  
\subsubsection*{Results}

Compared to other recent state-of-the-art systems in
Table~\ref{table:result},
our model achieves new state-of-the-art performance in two splitting methods with
great  improvements. As to our best knowledge,
our model is the first one that exceeds the 48\% accuracy in 11-way classification.

\begin{table}[ht]
  \centering
  \begin{tabu}{p{170pt}p{70pt}<{\centering}p{50pt}<{\centering}}
  \tabucline[0.65pt]{-}
  Model & PDTB-Lin & PDTB-Ji \\ \hline
  \newcite{lin-kan-ng:2009:EMNLP}
  & 40.20 & - \\
  \newcite{lin-kan-ng:2009:EMNLP}+Brown clusters
  & - & 40.66 \\
  \newcite{TACL536}
  & - & 44.59 \\
  \newcite{qin-zhang-zhao:2016:COLING}
  & 43.81 & 45.04 \\
  \newcite{qin-EtAl:2017:Long}
  & 44.65 & 46.23 \\ \hline
  Ours
  & \textbf{45.73} & \textbf{48.22} \\ \tabucline[0.65pt]{-}
  \end{tabu}
  \caption{Accuracy~(\%) comparison with others' results on PDTB 2.0 test set
  for 11-way classification.}
  \label{table:result}
\end{table}

\subsubsection*{Analysis}

\textbf{Ablation Study}

To illustrate the effectiveness of our model and the contribution of each module,
we use the PTDB-Ji splitting method to do a group of experiments.
For the baseline model,
we use 4 layer stacked convolutional encoder blocks without the residual connection in
the block with only pre-trained word embeddings.
We only use the output of the last layer and the output is processed by 2-max pooling
without attention and sent to the relation classifier and connective classifier.
Without the two residual connections, using 4 layers may be not the best for baseline
model but is more convenient to comparison.

Firstly, we add modules from high level to low level accumulatively
to observe the performance improvement.
Table~\ref{table:acc} is the results,
which demonstrate that every module has
considerable effect on the performance.

Then we test the effects of the two residual connections on the performance.
The results are in Table~\ref{table:res}.
The baseline$^+$ means baseline + bi-attention, i.e.,
the second row of Table~\ref{table:acc}.
We find that Res 1~(residual connection in the block) is much more useful than
Res 2~(residual connection for pair representation),
and they work together can bring even better performance.

\begin{table}[ht]
  \centering
  \begin{minipage}[t]{0.48\textwidth}
  \centering
  \begin{tabu}{ccc}
    \tabucline[0.65pt]{-}
  Level & Model & Performance \\ \hline
  - & baseline & 41.48 \\ \tabucline[0.1pt on 2pt off 2pt]{-}
  pair & + bi-attention & 42.25 \\
  sentence & + Res & 46.29 \\
  word & + subword & 47.03 \\
  word & + ELMo & 48.22 \\  \tabucline[0.65pt]{-}
  \end{tabu}
  \caption{Accumulatively performance test result.}
  \label{table:acc}
  \end{minipage}
  \hfill
  \begin{minipage}[t]{0.48\textwidth}
  \centering
  \begin{tabu}{cc}
    \tabucline[0.65pt]{-}
  Model & Performance \\ \hline
  baseline$^+$ & 42.25 \\ \tabucline[0.1pt on 2pt off 2pt]{-}
  baseline$^+$ + Res 1 & 45.33 \\
  baseline$^+$ + Res 2 & 43.57 \\
  baseline$^+$ + Res 1 + Res 2 & 46.29 \\  
    \tabucline[0.65pt]{-}
  \end{tabu}
  \caption{The effects of residual connections.}
  \label{table:res}
\end{minipage}
\end{table}

Without ELMo (the same setting as $4$-th row in Table~\ref{table:acc}),
our data settings is the
same as \newcite{qin-EtAl:2017:Long} whose performance was state-of-the-art
and will be compared directly. We see that even without the pre-trained ELMo encoder,
our performance is better, which is mostly attributed to our better
sentence pair representations.

\textbf{Subword-Level Embedding}
For the usefulness of subword-level embedding, we compare its performance to
a model with character-level embedding, which was ever used in
\newcite{qin-zhang-zhao:2016:COLING}.
We use the same model setting as the $4$-th row of
Table~\ref{table:acc}, and then replace subword with character sequence.
The subword embedding augmented result is \textbf{47.03\%},
while the character embedding result is \textbf{46.37\%},
which verifies that the former is a better input representation for the task.



\textbf{Parameters for Sentence-Level Module}
As previously discussed, argument specific parameter settings may result in better
sentence-level encoders.
We use the model which is the same as the third row in Table~\ref{table:acc}.
If shared parameters are used, the result is \textbf{45.97\%}, which is lower than
argument specific parameter settings~(\textbf{46.29\%}).
The comparison shows argument specific parameter settings indeed capture the
difference of argument representations and facilitate the sentence pair representation.

\begin{figure}[ht]
  \centering
  \includegraphics[width=0.5\textwidth]{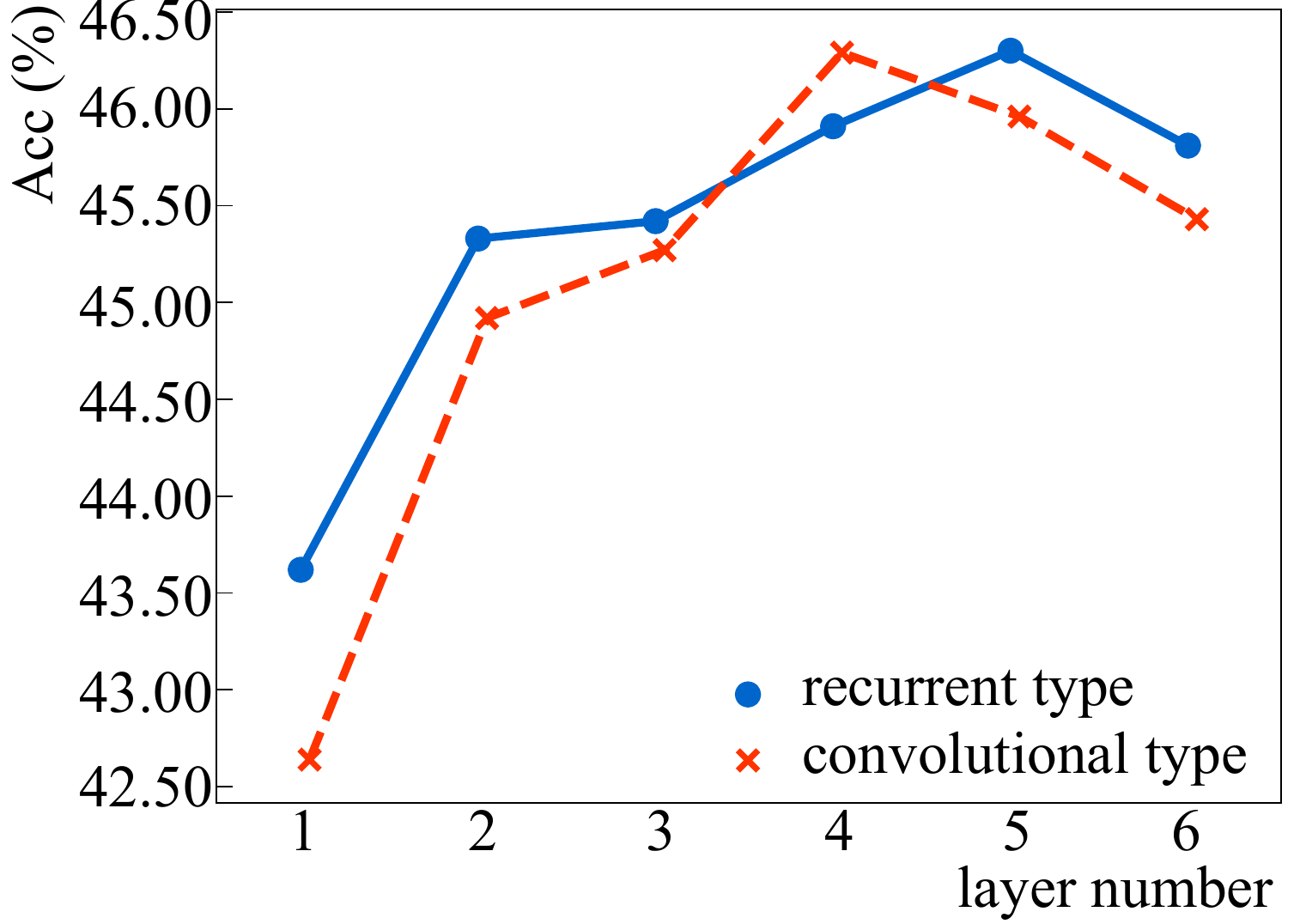}
  \caption{Effects of block type and layer number.}
  \label{fig:block}
\end{figure}

 
\textbf{Encoder Block Type and Layer Number}
In section 3.3, we consider two encoder types, here we compare their effects on the
model performance. Like the previous part,
The model setting is also the same as the third row in Table~\ref{table:acc}
except for the block type and layer number.
The results are shown in Figure~\ref{fig:block}.

The results in the figure show that both types may reach similar level of
top accuracies, as the order of word is not important to the task.
We also try to add position information to the convolutional type encoder,
and receive a dropped accuracy.
This further verifies the order information does not matter too much for the task.
For most of the other numbers of layers, the recurrent type shows better,
as the number of layers has an impact on the window size of convolutional encoders.
When convolutional type is used, the training procedure is much faster, but
choosing the suitable kernel size needs extra efforts.

\textbf{Bi-Attention}

\begin{figure}[ht]
  \centering
  \includegraphics[width=0.9\textwidth]{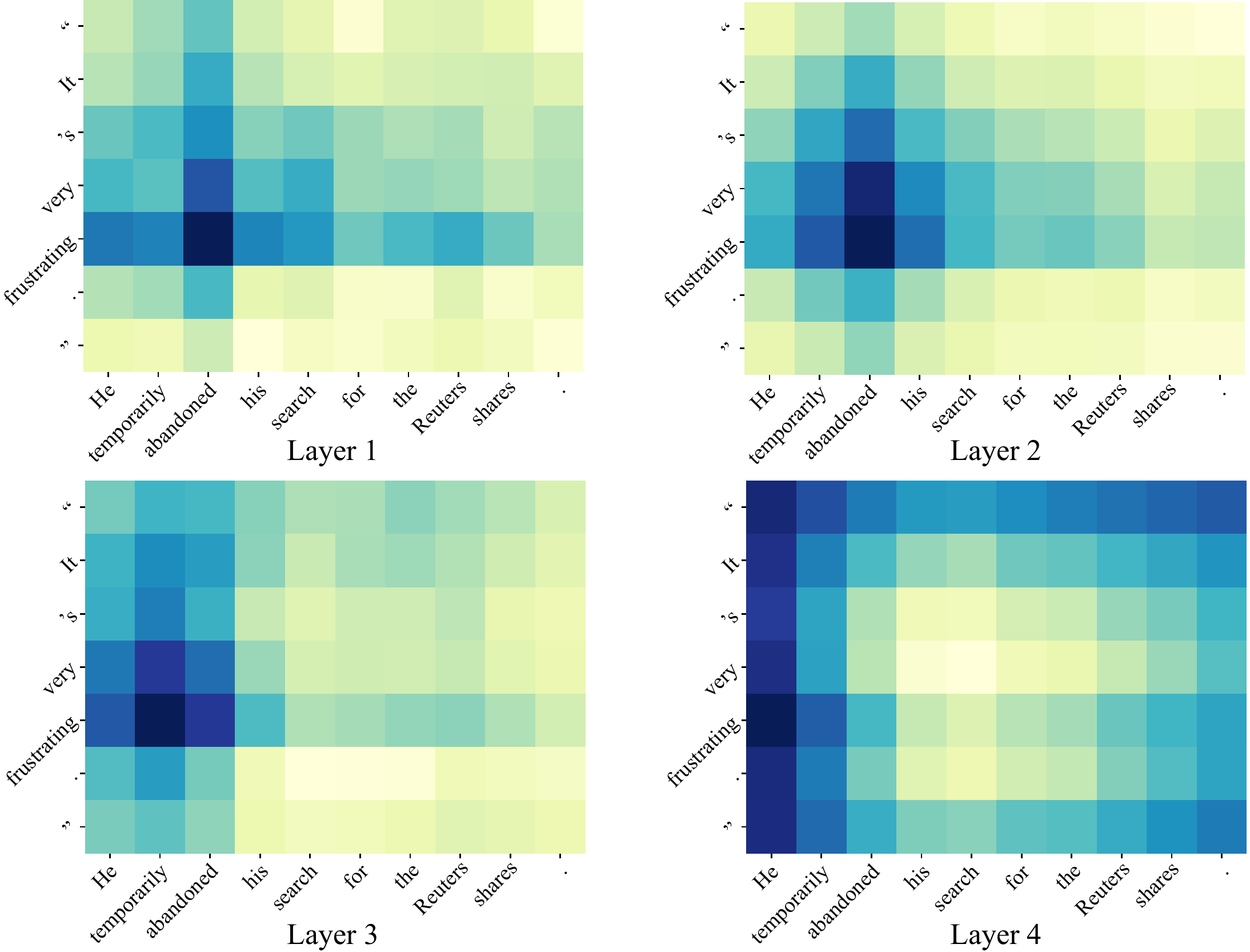}
  \caption{Attention visualization.}
  \label{fig:attn}
\end{figure}

We visualize the attention weight of one instance in Figure~\ref{fig:attn}.
For lower layers, the attended part is more concentrated. For higher layers,
the weights are more average and the attended part moves to the sentence border.
This is because the window size is bigger for higher layers,
and the convolutional kernel may have higher weights on words at the window edge. 

\subsection{Binary and 4-way Classification}

\textbf{Settings}
For the first level classification, we perform both 4-way classification
and one-vs-others binary classification.
Following the settings of previous works, the dataset splitting method is the same as
PDTB-Ji without removing instances.
The model uses 5 block layers with kernel size 3, other details are the same as that for 11-way
classification on PDTB-Ji.

\textbf{Results}
Table~\ref{table:result2} is the result comparison on first level classification.
For binary classification, the result is computed by $F_1$ score~(\%),
and for 4-way classification, the result is computed by macro average $F_1$ score~(\%).
Our model gives the state-of-the-art performance
for 4-way classification by providing an $F_1$ score greater than 50\%
for the first time according to our best knowledge.

\begin{table}[ht]
  \centering
  \begin{tabu}{p{130pt}ccccc}
  \tabucline[0.65pt]{-}
  Model & Comp. & Cont. & Exp. & Temp. & 4-way \\ \hline
  \newcite{zhang-EtAl:2015:EMNLP4}
  & 33.22 & 52.04 & 69.59 & 30.54 & - \\
  \newcite{TACL536}
  & 35.93 & 52.78 & - & 27.63 & - \\
  \newcite{chen-EtAl:2016:P16-13}
  & 40.17 & 54.76 & - & 31.32 & - \\
  \newcite{qin-zhang-zhao:2016:EMNLP2016}
  & 41.55 & 57.32 & 71.50 & 35.43 & - \\
  \newcite{liu-li:2016:EMNLP2016}
  & 39.86 & 54.48 & 70.43 & \textbf{38.84} & 46.29 \\
  \newcite{qin-EtAl:2017:Long}
  & 40.87 & 54.56 & 72.38 & 36.20 & - \\
  \newcite{lan-EtAl:2017:EMNLP20172}
  & 40.73 & \textbf{58.96} & 72.47 & 38.50 & 47.80 \\
  \newcite{lei2018linguistic}
  & 43.24 & 57.82 & \textbf{72.88} & 29.10 & 47.15 \\  \hline
  Ours
  & \textbf{47.85} & 54.47 & 70.60 & 36.87 & \textbf{51.06} \\  \tabucline[0.65pt]{-}
  \end{tabu}
  \caption{$F_1$ score~(\%) comparison on binary and 4-way classification.}
  \label{table:result2}
\end{table}

\section{Conclusion}

In this paper, we propose a deeper neural model augmented by different grained
text representations for implicit discourse relation recognition.
These different module levels work together and produce task-related
representations of the sentence pair.
Our experiments show that the model is effective and achieve the state-of-the-art
performance. As to our best knowledge,
this is the first time that an implicit discourse relation classifier gives
an accuracy higher than 48\% for 11-way and an $F_1$ score higher than 50\%
for 4-way classification tasks.

\bibliography{ref}
\bibliographystyle{acl}

\end{document}